# Mining Points of Interest via Address Embeddings: An Unsupervised Approach

ABHINAV GANESAN, Bundl Technologies (Swiggy), India
ANUBHAV GUPTA*, University of Maryland, United States
JOSE MATHEW, Bundl Technologies (Swiggy), India

Digital maps are commonly used across the globe for exploring places that users are interested in, commonly referred to as points of interest (PoI). In online food delivery platforms, PoIs could represent any major private compounds where customers could order from such as hospitals, residential complexes, office complexes, educational institutes and hostels. In this work, we propose an end-to-end unsupervised system design for obtaining polygon representations of PoIs (PoI polygons) from address locations and address texts. We preprocess the address texts using locality names and generate embeddings for the address texts using a deep learning-based architecture, viz. RoBERTa, trained on our internal address dataset. The PoI candidates are identified by jointly clustering the anonymised customer phone GPS locations (obtained during address onboarding) and the embeddings of the address texts. The final list of PoI polygons is obtained from these PoI candidates using novel post-processing steps that involve density-based cluster refinement and graph-based technique for cluster merging. This algorithm identified 74.8 % more PoIs than those obtained using the Mummidi-Krumm baseline algorithm run on our internal dataset. We use area-based precision and recall metrics to evaluate the performance of the algorithm. The proposed algorithm achieves a median area precision of 98 %, a median area recall of 8 %, and a median F-score of 0.15. In order to improve the recall of the algorithmic polygons, we post-process them using building footprint polygons from the OpenStreetMap (OSM) database. The post-processing algorithm involves reshaping the algorithmic polygon using intersecting polygons and closed private roads from the OSM database, and accounting for intersection with public roads on the OSM database. We achieve a median area recall of 70 %, a median area precision of 69 %, and a median F-score of 0.69 on these post-processed polygons. The ground truth polygons for the evaluation of the metrics were obtained using manual validation of the algorithmic polygons obtained from the Mummidi-Krumm baseline approach. These polygons are not used to train the proposed algorithm pipeline, and hence, the algorithm is unsupervised.

## 1 INTRODUCTION

A "point of interest" (PoI) is defined as a geographical location with simple associated metadata that could include a name, unique address identifier, information about the building at the location such as opening and closing hours, and other complex metadata like three dimensional model of the building at the location [28]. PoIs include wide categories of public and private spaces such as hospitals, shopping malls, restaurants, retail stores, residential compounds, business establishments, educational institutions, sports centres and parks. These PoIs are typically represented as a point on a digital map or as a polygon indicating the boundary of the PoI. We refer to polygon representations of PoIs as PoI polygons[1]. These representations are generally useful for people to either explore the locations of places they are interested in or navigate to these places.

Vast repositories of open-source data on PoIs are available as Voluntary Geographic Information (VGI). VGI includes crowd-sourced digital maps such as the OpenStreetMap (OSM) [13] and geotagged photos on Flickr. The use of raw VGI data for enterprise applications is limited by the absence of data quality guarantees. For example, data quality issues can arise due to various reasons such as lack of expertise of the volunteer in consistent use of terminologies (such as 'road' or 'street') [6] and absence of well-defined guidelines in countries like India where

---

*This work was done when the author was at Bundl Technologies.
[1]These are also known by other names like areas or regions of interest.

Authors' addresses: Abhinav Ganesan, g.abhinav@swiggy.in, Bundl Technologies (Swiggy), P.O. Box 1212, Bangalore, Karnataka, India; Anubhav Gupta, anubhav@umd.edu, University of Maryland, College Park, Maryland, United States; Jose Mathew, jose.matthew@swiggy.in, Bundl Technologies (Swiggy), Bangalore, Karnataka, India.

addresses can be diversely structured. VGI however is still useful for extracting some intelligence attributes for maps such as PoIs.

Ahern et al. [1] used K-means clustering [5] to cluster the locations on geotagged photos from Flickr, merged the clusters using heuristics based on distance, and shortlisted clusters where the products of term-purity (TP)[2] and suitably redefined Term Frequency-Inverse Document Frequency (TF-IDF) [24] of the tags exceeded a threshold. The TP factor measures the fraction of points within a cluster that contains a certain tag. The TF-IDF criterion measures the fraction of points within a cluster that contains a tag as opposed to the appearance of the tag in the overall corpus. The work by Mummidi and Krumm [22] used text annotated geographical pushpins that were obtained by crowdsourcing for an early version of Microsoft Bing maps. This work used hierarchical agglomerative clustering [5] on the geographical locations and independent thresholds on TF-IDF and TP of $n$-grams[3] for each cluster in order to shortlist the PoIs. The extracted $n$-grams were declared to be the PoI names.

The evaluation of the algorithmic output in [1] was qualitative, and in [22], it was semi-quantitative. In [1], a digital user interface (UI) namely, the World Explorer was built and participants were recruited to evaluate the extracted PoI tags. The evaluation was descriptive and oriented towards tourists' perspectives of exploring familiar and unfamiliar geographies in a city using the UI. In [22], the evaluation was done by volunteers who marked if the extracted PoI location in a small geography of the city of Seattle was recognisable either through the extracted name or the approximate location. Approximately 80% of the PoIs identified were not found to be recognised by the users. The algorithmic output was divided into "control" and "test" based on whether the PoI was found in a Yellow Pages database or not. Among the test outputs, the PoIs were deemed correct in their approximate locations and names 76.8 % of the times. For the control outputs, the PoIs were deemed correct 92.2 % of the times.

Mining PoI polygons using keyword search in Flickr's geotags was explored in [4] and evaluated using area-wise precision and recall (defined in Section 5, proposed in [9]) for 24 PoIs each in Rome and Paris. For instance, geotags with the keyword *Colosseum* with a fixed set of spell-variants in European languages were chosen. The locations of these geotags were iteratively filtered out using a density criteria and the PoI polygon is obtained as a convex hull of the remaining set of points at convergence. In [9], PoI polygons were mined from the Yellow pages dataset by correcting the locations using a geocoding service and using Voronoi tessellations of OSM polygons. The work however assumes that every location in the Yellow pages dataset corresponds to a distinct PoI since the dataset is a business directory that lists local businesses.

## 2 MOTIVATION AND CONTRIBUTIONS

Our work focuses on mining PoI polygons in the context of online food ordering and delivery platforms. Online food deliveries happen in a hyperlocal setting with bounded delivery times, typically under thirty minutes to deliver an order. Customers onboard their addresses using their smartphones indicating the location where they want the delivery to happen. Each address for every customer is assigned a unique identifier *address_id*. Each *address_id* is associated with an address text, which is the textual representation of an address, and an address location which is the geographical location of the address collected using GPS and represented as a (latitude (lat), longitude (lng)) pair. Examples of address texts are given in Table 1 in page 6. The PoIs, which are not necessarily commonly visited public spaces, need to be identified from such onboarded address texts and address locations. To reiterate, these PoIs could represent residential complexes, office spaces, hostels, and hospitals to name a few. At enterprise scale of data, these PoIs could be densely located in a given geography, and hence, algorithmically locating these PoIs accurately becomes challenging even with location data. Our dataset, use-case, and hence, the structure of the address data is very different from the datasets used in the works described in the previous section. Some of the challenges in our dataset are listed below.

---

[2]The terminology of term purity was first used in [22] which was published at a later date than [1]. However, we use the same terminology for simplicity since the criterion used in [1] was conceptually similar to TP used in [22] except for considering unique photographers who included the tag in the context of Flickr dataset.

[3]An $n$-gram is defined as a sequence of $n$ consecutive words [19].

- Customer addresses in India are not structured homogeneously like in some developed western nations. In other words, the customer addresses are not neatly partitioned into their entities such as flat number and the PoI name. It's practically impossible to enforce this at the scale we operate.
- Customers do not use consistent spellings in English. A vast majority of the PoI or locality names written in English are rooted in Indic languages which are mostly phonetic while English is not a phonetic language.
- It's not known a priori if an address belongs to a PoI. For instance, the address in the first row of Table 1 belongs to a PoI (a residential apartment) while the addresses in the other rows do not belong to PoIs, but this is not known until we manually investigate the addresses. This is not the case in [9] where it is known that the addresses in the dataset belong to PoIs. The PoI names are also unknown a priori. This is not the case in [4] which mines PoIs using PoI names for keyword searches in Flickr's geotags.

The challenge lies in mining PoIs from unstructured addresses that are written with diverse spell variations in dense urban settings where traditional approaches based on location-only clustering or using Term Purity like in [22] have limitations. Location-only clustering based on density has no address text intelligence and are hard to tune in Indian settings where the addresses have high variance in the geographical distribution. The approach in [22] does not account for spell variations. For instance a PoI by name *HappiStay* is written as *Happy Stay*, *Happystay*, and *Happi stay*. These spell variations are best addressed through contextual text embeddings (summarized in the next section).

Some use-cases that the PoI polygons serve in our context are listed below.

(1) These boundaries serve us in batching orders for deliveries for those originating from the same PoI around the same time. Batched orders are picked up and delivered by the same delivery executive (DE) in a sequence with no other orders picked up or delivered in between them by the DE to ensure efficient deliveries.
(2) Knowledge of entry gates to the PoIs helps with efficient last-mile routing. In other words, it helps to route our delivery partners to the closest entry gate for a customer order in case of large PoIs with multiple entry gates. The PoI boundary helps in algorithmically identifying entry gates to the PoIs by overlaying our Delivery Partners' (DE) GPS trajectories collected during order deliveries. Note that we are not interested in building footprints themselves which are the typically available polygons on the OSM database. We want to draw a boundary that encompasses the whole private access compound of a residential apartment or an office building. Such boundaries are useful in identifying entry gates to the PoIs.
(3) Knowledge of PoI boundary helps segment customers which in turn can be used for targeted coupons and discounts.

The contributions and organisation of this work are as follows.

- The address texts are preprocessed using locality data obtained from real estate websites in India. These address texts are converted into a 300-dimensional contextual vectorial representation (Add2Vec) using a deep learning-based model, RoBERTa [17]. We concatenate this address embedding with the address location given by (lat, lng) using a scaling factor to form a feature vector for every address id. This feature vector is used for hierarchical agglomerative clustering to identify PoI candidates.
- We propose a cluster homogeneity criterion to shortlist the clusters as PoI candidates. To the best of our knowledge, this is the first work that uses locality name-based pre-processing and joint clustering of location and address embeddings to mine PoIs. The details appear in Section 4.3 and Section 4.5.1.

- The shortlisted PoI candidates are then post-processed in a novel manner by eliminating noisy GPS points in the clusters, merging clusters with similar names through a graph-based technique, discarding redundant polygons, and merging intersecting polygons. The details appear in Section 4.5.2-4.5.4.
- The metrics achieved by the proposed algorithm are listed in Table 3 in Section 5 evaluated across nine major cities in India. This is the first reported unsupervised algorithm for mining PoIs run on a pan-India scale with published metrics. Our algorithm also identified 74.7% more PoIs than that identified by the Mummidi-Krumm baseline algorithm in [22] run with relaxed parameters over our internal dataset.
- In order to improve the median F-score of the algorithmic polygons, we propose polygon boundary correction using building footprint polygons and roads data in the OSM database in Section 6. This algorithm makes use of OSM polygons that intersect with the algorithmic polygon and expands the boundary of the polygons if and when a "closed private road" encompasses them. The algorithm also retains only the largest polygon if the resulting polygon cuts across a public road. The metrics achieved using OSM based polygon correction are listed in Table 3. The algorithm can be retrofitted to any PoI mining algorithm that mines PoI polygons using address location and text.

## 3 OTHER RELATED WORKS

In the context of mining intelligence from Indian addresses for e-commerce applications, Ravindra et al. have dealt with problems such as operational zone classification of addresses for last-mile deliveries [2, 18], classification for randomly typed addresses [3], and address clustering based on the similarity between addresses [14]. The clustering algorithms for addresses proposed in [14] were not targeted at any particular task such as mining PoIs. The clusters obtained were evaluated using Silhouette coefficient [25] and Calinski-Harabasz index [7] for clusters obtained through the leader clustering algorithm with cosine similarity metric over Word2Vec based [21] address embeddings. These metrics when high are indicative of dense and well-separated clusters in the vector space of address embeddings. However, these metrics do not convey how good the clustering algorithm is for mining PoIs. It is possible that addresses within the same locality or street happen to be well-formed clusters as indicated by these metrics because of the fact that such addresses could look similar. Nevertheless, such metrics are not useful from the point of view of the goodness of PoIs, especially in dense geographies. The goodness of a PoI mining algorithm can instead be directly measured using the area-wise precision and recall metrics described in Section 5.

In [18], the authors used state-of-the-art deep-learning-based architecture called RoBERTa [17] for sub-zones classification task and showed that this architecture outperforms other architectures for classification based on Word2Vec [21] and Bi-LSTM [20] for embedding the addresses. The sub-zones represent geographical regions where the shipments meant for customers are batched at some stage in the last-mile delivery chain in an e-commerce setting. We, therefore, use RoBERTa as the architecture of choice in our work. However, note that the goal of our work is not classification. The authors also proposed novel pre-processing steps for the address texts which we collectively refer to as *vocabulary pre-preprocessing*. We adopt these pre-processing steps with minor modifications. In the next section, we describe our algorithm for mining PoIs.

## 4 MINING POI POLYGONS: SYSTEM DESIGN

The end-to-end system design is presented in Fig. 1. Each address is identified by its location and address text. The modules in the system are explained in the subsequent sub-sections.

### 4.1 Vocabulary Pre-processing

This address text pre-processing step adopts the same steps for preprocessing as listed in [18] with minor modifications. We call this vocabulary pre-processing and use this to "standardise" the addresses. Similar to [18], we partition the address corpus of approximately 6 million addresses across India into five geographical zones, namely south, central, north, east, and west. The following pre-processing steps are applied to the address texts within the zones.

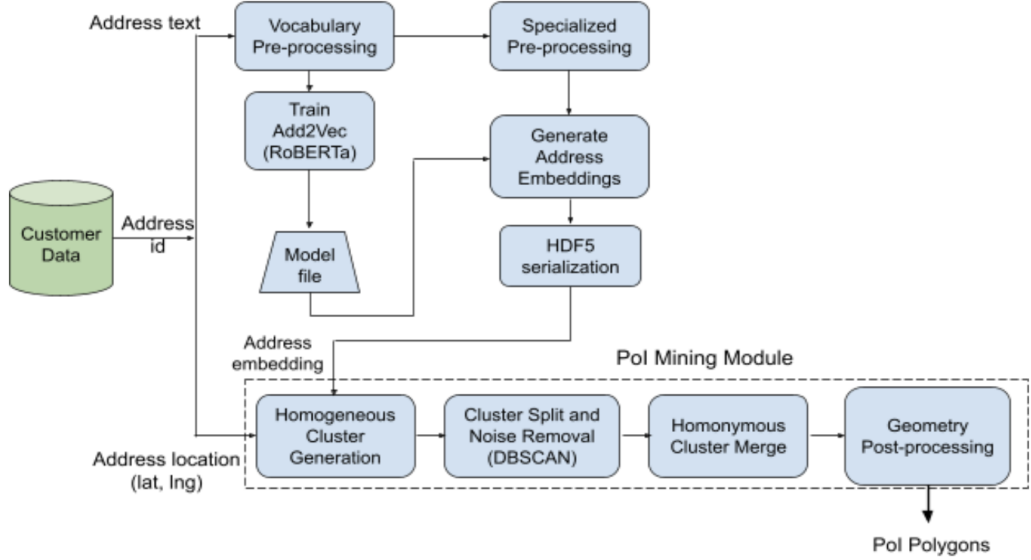

Fig. 1. System Design for mining PoI polygons from address texts and locations.

(1) *Common text pre-processing:* This step comprises of substituting punctuations and special characters by a single space, replacing whitespaces by a single space, and lower-casing all the alphabets.
(2) *REGEX split:* Split alphanumeric words into words containing the only alphabet and numeric characters. This split can be done using a regular expression (REGEX) pattern search. Customers often miss out on including space between the alphabet and numeric characters. These numeric characters could represent floor numbers, flat numbers, or Postal Index Number pin-codes. This step was not a part of preprocessing in [18]. An example of this from our address corpus is the city name and pin-code joined together as *bangalore560066* which is split as *bangalore 560066*.
(3) *Probabilistic word split:* Here, the words are split based on empirical probability of occurrence. A word is split into two at any location if the product of probabilities of the individual words is more than the probability of the compound word. Here, we use the empirical probability of the words as given in [2] instead of word count as given in [18] since we found that empirical probability-based splits gave qualitatively better word splits. An example of this split in our corpus is *borewellroad* split into *borewell road*.
(4) *Bigram separation:* The probabilistic word splitting step does not account for spell variations that one encounters in the Indian context. If a split word has insufficient support in the corpus but happens to be a spelling variant of a valid locality or a PoI name, such word splits might get missed. Therefore, the authors of [18] propose to construct a dictionary of bigrams and those words which are close to these bigrams in the sense of edit distance [16] and phonetic match[4] [23] are split into these bigrams. For example, in our corpus, the mistyped word *Jaibheemanagr* was replaced by the bigram *jaibheema nagar*.

Table 1. Examples of vocabulary pre-processed addresses. The words that underwent one of the vocabulary pre-processing steps (3)-(5) are highlighted in bold. The flat number identifiers in the addresses are anonymised as XXX.

| Preprocessing step | Raw address | Vocabulary pre-processed address |
|---|---|---|
| Probabilistic word split | flat No XXX,**srivenkateswara** Nilayam. Kukatpally. Hyderabad 500072 | flat no XXX **sri venkateswara** nilayam kukatpally hyderabad 500072 |
| Bigram Separation | No XXX N Block 18 th Street East **Annanager** Chennai | no XXX n block 18 th street east **anna nagar** chennai |
| Probabilistic word merge | XXX, Padmaja Nagar , Vemana Colony, **Chanda Nagar**, Hyderabad, Telangana 500050, India | XXX padmaja nagar vemana colony **chandanagar** hyderabad telangana 500050 india |

(5) *Probabilistic word merge:* This step is similar to probabilistic word splitting. This is required because customers inadvertently include whitespaces within words when the regular spelling does not include them. An example of this merging is the bigram *ram krishna* merged to be *ramkrishna*.

Note that we do not include the spell-correction step from [18] that involves edit distance and phonetic match between the word pairs since we observed that it resulted in many false corrections qualitatively. These false corrections were more pronounced in the case of unigrams and much less in the case of bigram separation. For example, the word *bhuvana* got corrected to *bhavana*. These are completely different words and phonetic match doesn't capture the pronunciation difference between the words well in the Indian context. A few examples of vocabulary pre-processed addresses appear in Table 1.

### 4.2 Add2Vec Model training

The zone-wise vocabulary preprocessed address dataset is used to train a masked language model (MLM) to learn the address embeddings. We use the RoBERTa architecture [17][5] which was demonstrated to produce reasonable contextual embeddings for a classification task in [18] for Indian addresses. No train-test data split is used here since the focus of this work is not on generalisable models on unseen data. The address texts are tokenised using Byte Pair Encoding (BPE) [12] which comes packaged in a standard HuggingFace implementation [29]. The address token sequences in a batch are padded to the maximum length of the token sequences in the batch or truncated to a length of *max_position_embeddings*. The hyperparameters used are given below.

$$vocab\_size = 52000, max\_position\_embeddings = 70,$$
$$num\_attention\_heads = 10, num\_hidden\_size = 300,$$
$$num\_hidden\_layers = 6, batch\_size = 256, num\_epochs = 10.$$

The address embeddings are generated from the last layer of dimension *num_hidden_size*. The default values for *num_hidden_size* and *num_attention_heads* are equal to 768 and 12 respectively [17]. However, the default dimension of 768 makes the subsequent stages of processing infeasible in terms of run-time for mining PoIs. Therefore, we use a reduced dimension of 300, and since, the architecture requires that the *num_attention_heads* needs to be a factor of *num_hidden_size*, we use *num_attention_heads* = 10. The value chosen for *max_position_embeddings* is similar to the one used in [18] since our addresses are of similar length.

---
[4]Two words are said to phonetically match if they sound similar. Algorithms for phonetic matching do not necessarily perform well in the context of Indian addresses.
[5]Description of the architecture is out of scope of this paper and we refer the readers to the original paper for the same.

Table 2. This table illustrates examples of specialised pre-processing of addresses. In the first example, the unigrams 'floor', 'street' and 'chennai' are eliminated due to top-words filtering, and the shortened pincode '28' is removed due to character pre-processing. The unigram 'mandaveli' indicates a locality name that is not removed because it is not present in our locality list. In the second example, the (sub)locality unigrams 'indiranagar' and 'adyar' are eliminated during the locality pre-processing step, and the unigrams 'st' and 'chennai' are removed using the top-words pre-processing step. The anonymised numbers XX are eliminated in the character pre-processing step.

| Vocabulary pre-processed address | Specialized pre-processed address |
|---|---|
| XX navins lakshmi ram apartment ground floor thiruvengadam street mandaveli chennai 28 | navins lakshmi ram apartment ground thiruvengadam mandaveli |
| XX sudsun shevroy apts 1 st cross street indiranagar adyar chennai 600020 | sudsun shevroy apts cross |

### 4.3 Specialised Address Pre-processing

The RoBERTa model trained on the vocabulary pre-processed addresses is inferenced on the addresses which undergo specialised pre-processing for mining PoIs. For online food delivery, the trained model is also useful for other use-cases such as fraud detection and address anomaly identification. For horizontal scalability in an industrial setting, we do not train the model over address texts that go through the complete pre-processing pipeline specialised for mining PoIs. Instead, we train them on "standardised" address texts where the standardisation is achieved through vocabulary pre-processing.

The specialised preprocessing for address texts involve the following steps.
  (1) *Character pre-processing:* This step comprises of removing numeric characters and removing single alphabets. Removing numeric characters helps remove pin-codes, flat numbers, and floor numbers that do not carry any information on the PoI names. Similarly, single alphabets rarely carry information on the PoI names, and hence, are removed.
  (2) *Top-words and locality removal:* Here, we remove top-10 frequently occurring unigrams and (sub)locality names from the address texts city-wise. The (sub)locality names are tokenised into unigrams and the unigrams are used as stop-words in the pre-processing. In the city of Chennai in India, the top-10 frequently occurring unigrams are 'street', 'chennai', 'nagar', 'road', 'block', 'th', 'st', 'floor', 'flat', 'no'. The city-wise locality names are obtained from real-estate websites, but these are not an exhaustive list. Not including this pre-processing step resulted in 23% false polygons that represented localities even in the Mummidi-Krumm baseline algorithm evaluated in a small geography in a top Indian city.

Examples of specialised pre-processed addresses are given in Table 2.

### 4.4 Generating Address Embeddings

The specially preprocessed addresses are inferenced through the trained RoBERTa model and the address embeddings are generated as the mean of the hidden states of the last layer (a fully connected layer) corresponding to the input tokens of the addresses. These address embeddings are serialised in an HDF5 format [27] to be consumed by the subsequent clustering step. The file can be accessed in a similar manner as arrays while the serialised data is not stored on the RAM. This is essential because the PoIs need to be mined from a large dataset of millions of addresses.

### 4.5 PoI Polygon Mining

The PoI polygon mining is a 4-step process as highlighted under 'PoI Mining Module' in Fig. 1. These steps are run parallely on addresses whose locations are partitioned into L5 geohashes [26] and subsequently into geographical bins. The geographical bins are generated as 5x5 linearly spaced partitions between the minimum and maximum latitude and longitude of the customer locations present in the geohash so that each bin is roughly 1 km$^2$. The PoI mining module is described in the subsequent sub-sections and Fig. 2 shows the algorithm in action on the page after the next. Some of the largest residential and office PoIs are approximately of area 500 m$^2$ in India, but these are relatively too few in number. So, the chosen bin size ensures that with high probability the PoIs fall within the bins. All the parameters used in this section were chosen by running the PoI mining module over approximately 9000 addresses in a bin in a large Indian city and qualitatively validating the polygons obtained. We could run the algorithm at scale (for 6 million addresses as mentioned in Section 4.1) on a shared Spark cluster in 45 minutes.

*4.5.1 Generate homogeneous clusters.* The PoI candidates are generated as homogeneous clusters as follows.
  (1) Concatenate the normalised customer location ($lat_{norm_i}, lng_{norm_i}$) and address embedding $v_i$ generated as in Section 4.4 for every address id to be a feature vector $f_i = (\lambda lat_{norm_i}, \lambda lng_{norm_i}, v_i)$ with a location-scale of $\lambda = 10$, for $i = 1, 2, \cdots, N$ where $N$ represents the number of address ids in the geographical bin. The normalised locations ($lat_{norm_i}, lng_{norm_i}$) are mean 0, variance 1 normalised versions of the actual customer locations ($lat_i, lng_i$) within a geographical bin. The latitudes and the longitudes are normalised independently. We now perform hierarchical agglomerative single linkage clustering [5] on the feature vectors $f_i$. The location scale is chosen by qualitatively evaluating the outputs in a geographical bin. Note that if $\lambda$ is large, the clustering "tends" to the location-only clustering in [22], while if $\lambda$ is small, the address embedding proximity between the addresses dominates the clustering. In the former case, the similarity between addresses is not taken into account, and in the latter case, "falsely" close address embeddings in the Euclidean space tend to get clustered together. Only clusters containing at least 10 points are retained.
  (2) Homogeneous clusters are obtained as clusters where address embeddings of 90 % of the points in the clusters have a cosine distance of at least 0.9 with the median centroid address embedding of the cluster. The median centroid embedding of the cluster is defined to be the coordinate-wise median of the address embeddings of all the points within the cluster.
  (3) Redundant homogeneous children clusters are discarded, i.e., only homogeneous clusters at the top of the agglomeration are retained and all the children clusters that also cleared the homogeneity checks mentioned above are discarded.

*Note*: It is possible that addresses that don't belong to the same PoI pass the address homogeneity check because the address embeddings end up being similar due to common locality names or relatively common names like 'floor' or 'ground' despite choosing a high cosine similarity threshold. It also remains to be explored if training the Add2Vec itself on the preprocessed addresses does better in such cases.

*4.5.2 Splitting clusters and GPS noise clean up using DBSCAN.* The input into this step is the set of PoI candidate clusters from the previous sub-section and the customer locations that constitute these clusters. Each PoI candidate cluster is either discarded or split up into multiple candidate clusters or retained as it is at the end of this step of processing. The noisy GPS locations result in misleading PoI boundaries and this step cleans up the noise using density-based clustering namely DBSCAN [11]. This step also serves the dual purpose of splitting up the points that were falsely clustered together into a single PoI candidate cluster. DBSCAN uses a point density check-in finding at least *num_neighbours* = 5 address locations within a neighbourhood of $\epsilon = 10m$. If a customer location does not have sufficiently many neighbours, the location is discarded as noise. A cluster is grown by recursively scanning the points for dense neighbours and scanning the neighbours for further dense neighbours and so on. A new cluster is formed at a point that is not in the dense neighbourhood of the points in the earlier clusters. There

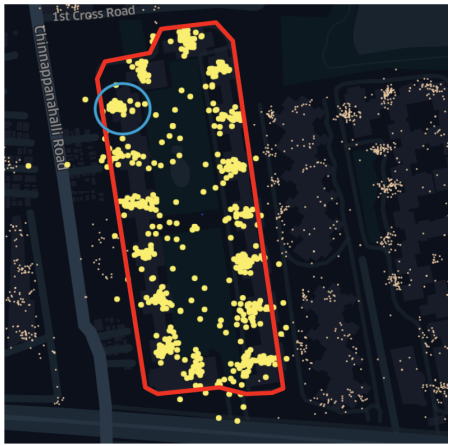
(a) Generating homogeneous clusters.

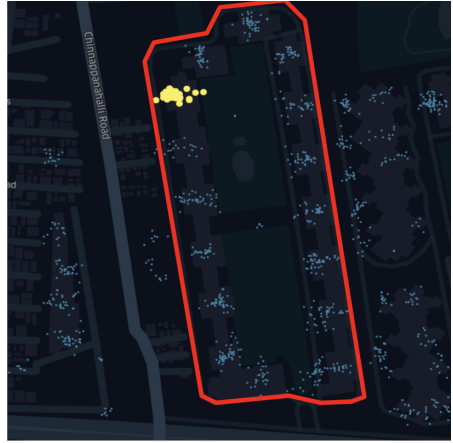
(b) DBSCAN on homogeneous clusters.

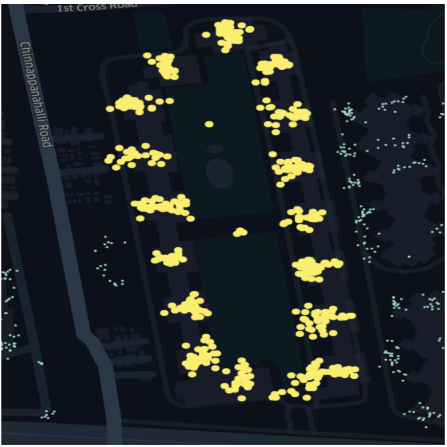
(c) Merging homonymous clusters.

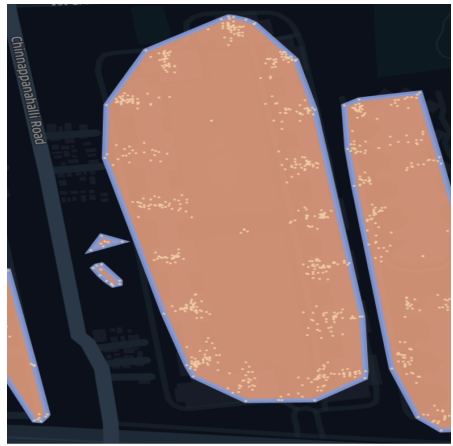
(d) Convex hull

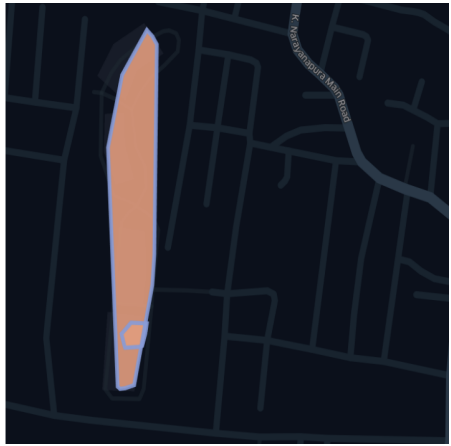
(e) Substantially Intersecting Polygons

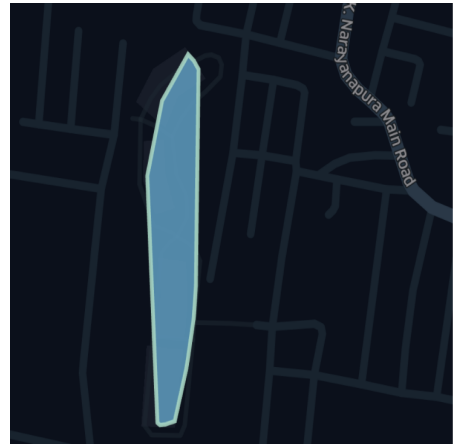
(f) Merge substantially intersecting polygons

Fig. 2. An illustration of the PoI mining algorithm. The red polygon indicates the ground truth PoI polygon. The highlighted yellow points in (a) indicates points within a single homogeneous cluster. The homogeneous cluster is split into multiple clusters (the blue points in the red polygon that appear clustered) by DBSCAN and noisy locations are discarded as shown in (b). The cluster highlighted in (b) is obtained from the blue circled set of points in (a). Homonymous cluster merge indicated by the points highlighted in yellow is shown in (c). The split clusters in (b) were identified to have the same high confidence names, and hence, got merged. The convex hull of the merged cluster is shown in (d). An example of a small polygon substantially intersecting with a big polygon appears in (e) and the convex hull of the two polygons is shown in (f).

are several prior works that use DBSCAN or its variants for extracting the PoIs themselves [15][6]. However, here we use DBSCAN as a post-processing step.

*4.5.3 Merging homonymous clusters.* Splitting the PoI candidate clusters in the previous step results in splitting some genuine sub-PoIs (like towers) within the same PoI (like residential complexes). This step generates high confidence names for PoI candidate clusters from the previous step, if such high confidence names can be extracted, and merges the clusters with similar high confidence names with minor spell variations which is captured by edit distance. The clusters with similar high confidence names are referred to as *homonymous clusters*. The steps involved are listed below.

(1) Generate n-grams for $n = 2, 3, 4$ from the list of addresses that are present in each PoI candidate cluster.
(2) Classify the clusters into ones with high confidence names and low confidence names, namely high confidence name clusters (HCNC), and low confidence name clusters (LCNC). High confidence names are the n-grams that are present in at least 70 % of the addresses in the cluster. There could be multiple high confidence names for a given cluster. The other clusters are classified as LCNC.
(3) For HCNCs, form a graph with the clusters as the nodes. Nodes have an edge between them iff there is a common high confidence name between the clusters up to an edit distance of one (e.g. 'mangalya suryodaya' and 'ma**a**ngalya suryodaya' ) and distance between the cluster centroids is less than a threshold of 100 m.
(4) Run depth-first search (DFS) [8] to identify connected components in the above graph. Merge the connected component HCNCs into a single cluster.
(5) The merged HCNCs, the HCNCs that are not merged, and the LCNCs are passed on to the next step.

*4.5.4 Post processing on PoI polygons.* This step comprises generating the polygon representation of the PoIs, discarding redundant polygons, and merging intersecting polygons. More specifically, the steps are listed below.

(1) The PoI polygons are generated as the convex hull of the locations contained within the clusters from the previous step.
(2) Embedded polygons are discarded, i.e., those which are completely contained within the others are discarded. Note that this is also partially achieved by discarding children clusters in Section 4.5.1. However, there might still be clusters that aren't necessarily parent clusters, but in the polygon representation, they might contain polygons from other clusters. This can happen due to differences in the way addresses are written and consequently don't get clustered in a parent-child hierarchy in the Euclidean space.
(3) Merge intersecting polygons if, at least one of the intersecting polygons have a 70 % area overlap with the other polygon. It might happen that there are multiple such intersecting pairs with shared polygon candidates. To handle these cases, this step requires a little more sophistication which is described below.
(4) Form a graph with the polygons as vertices. Edges are drawn between these vertices if they intersect and at least one of the intersecting polygons have a 70 % area overlap with the other polygon. We run DFS to get connected components in this graph. These connected components are merged by computing the convex hull of the connected components.

Note that we merge polygons only if they intersect substantially. This is because, in dense urban geographies in India where small buildings are close to each other, a significant number of noisy customer locations spill over to the adjacent buildings.

---

[6]We refer the reader to [15] for a comprehensive overview of DBSCAN based algorithms.

Table 3. Metrics of algorithmic polygons and polygons corrected using OSM data as compared to that for the baseline Mummidi-Krumm algorithm.

| Metrics | Mummidi-Krumm Baseline PoI Polygons | OSM Corrected Baseline PoI Polygons | PoI Polygons from Proposed Mining Algorithm | OSM Corrected Proposed Algorithmic Polygons |
|---|---|---|---|---|
| Median area precision ($P$) | 80.7% | 68% | 98.7% | 69% |
| Median area recall ($R$) | 9.6% | 85% | 8.2% | 70% |
| Median F-score $\frac{2}{\frac{1}{P}+\frac{1}{R}}$ | 0.17 | 0.76 | 0.15 | 0.69 |
| Number of PoIs | $x$ | $0.67x$ | $y = 1.74x$ | $0.55y$ |

## 5 EVALUATION OF THE ALGORITHM

We use the following area based evaluation metrics [9].

$$\text{Area Precision} = \frac{\text{Area}(\mathcal{P}_A \cap \mathcal{P}_G)}{Area(\mathcal{P}_A)},$$
$$\text{Area Recall} = \frac{\text{Area}(\mathcal{P}_A \cap \mathcal{P}_G)}{Area(\mathcal{P}_G)},$$

where $\mathcal{P}_A$ and $\mathcal{P}_G$ represent the proposed algorithmic and ground truth PoI polygons, and only the polygon pairs with a non-zero intersection area are used for the metric computation. We run the Mummidi-Krumm algorithm on our data with reduced thresholds of TP=0.7, TF-IDF= 0.1, and the number of points per cluster set to 15 to generate PoI polygons. These thresholds were relaxed compared to the thresholds TP=0.9 and TF-IDF= 0.9 used in [22][7] to enhance the coverage of polygons (i.e., the number of polygons). The polygons were post-processed to merge polygons with the same name since we used a low TF-IDF threshold. The boundaries of the polygons were manually corrected using satellite imagery to fit the actual boundary of the PoI. Also, only 72% of these polygons were found to represent valid PoIs. The rest of them turned out to be polygons containing locality names or street names. These manually validated polygons are used as ground truth[8] to evaluate the proposed algorithm in this paper. The proposed algorithm identified 74.8% more PoIs than that identified by the Mummidi-Krumm baseline algorithm and 67.5% of the ground truth polygons are identified by the proposed algorithm. The cumulative distribution function (CDF) of area precision and area recall of pairs of intersecting algorithmic and ground truth PoI polygons are presented in Fig. 4. The median area precision is 98.7% and the median area recall is 8.2%. This means that 50% of the intersecting algorithmic polygons lie well within the ground truth polygons and do not cover the ground truth polygons completely. For better understanding, examples of high precision and low recall, and low precision and high recall are presented in Fig. 3. The median F-score (defined in Table 3) is given by 0.15. These results along with the comparison against the metrics for the baseline polygons are summarized in Table 3. In summary, the median F-scores of the proposed algorithm and the baseline algorithm are similar, but the proposed algorithm identifies 74.8% more PoIs than the baseline algorithm.

---

[7]We reiterate that the work in [22] focussed on a specific sub-region of Seattle and are not evaluated on Indian addresses.
[8]The total number of these polygons is a few tens of thousands.

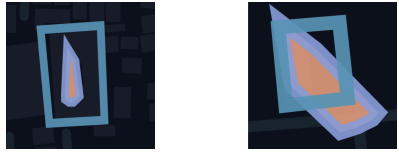

Fig. 3. The first example where the algorithmic polygon is contained well within the ground truth polygon represents a case of high precision and low recall. The second example where the algorithmic polygon substantially leaks out of the ground truth polygon, but substantially overlaps with the ground truth polygon represents a case of low precision and high recall.

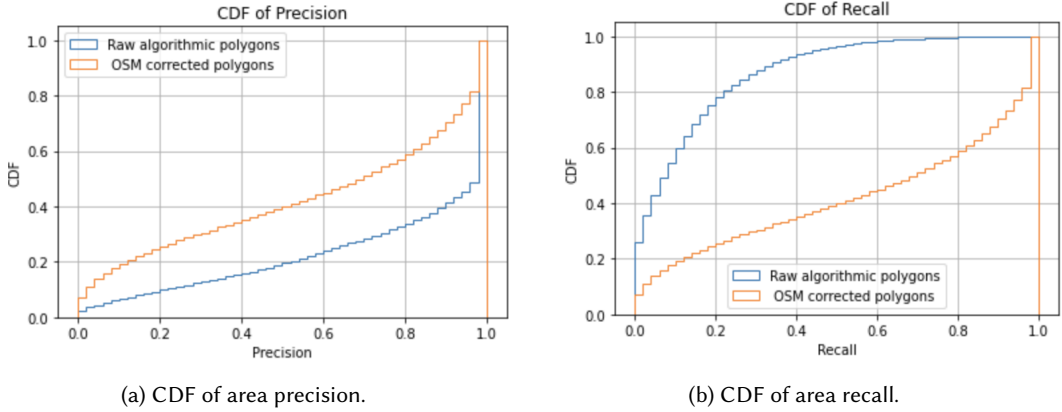

(a) CDF of area precision.

(b) CDF of area recall.

Fig. 4. A CDF value of $y$ represents the empirical probability of precision or recall values being at most a given value $x$, i.e., $y$=probability($X <= x$). The median value is obtained at $y = 0.5$. With OSM based polygon correction, the recall CDF curve shifts substantially to the right (indicating improvement) and the precision CDF curve shifts relatively marginally to the left (indicating degradation) compared to the CDF curve for raw algorithmic polygons.

## 6 POLYGON CORRECTION USING OSM BUILDING FOOTPRINTS AND ROADS

Improving the recall without degrading the precision of the polygons reduces the manual validation time. We only present the basic idea here through Algorithm 1 and 2 and briefly explain them below. A detailed flowchart along with examples of the algorithms in action is given in Appendix A.

---

**Algorithm 1** Pruning polygons using public roads

1: **procedure** PRUNE_POLYGON_VIA_HIGHWAY($polygon$)
2:     $d \leftarrow max(distance\ between\ 2\ points\ on\ polygon)$
3:     $roads \leftarrow$
        $OSM\ non\text{-}circular\ roads\ intersecting\ with\ polygon$
4:     **for** $road$ in $roads$ **do**
5:         $len \leftarrow length\ of\ road$
6:         **if** $road \in Public\ Roads$ **or** $len \geq 2 * d$ **then**
7:             $subPolygons \leftarrow split\ polygon\ into\ parts\ by\ road$
8:             $polygon \leftarrow largest\ sub\text{-}polygon$
9:     **return** $polygon$     ▷ The pruned polygon

**Algorithm 2** Correcting polygons using intersecting and encompassing private and public roads

1: **procedure** CORRECT_POLYGON_VIA_HIGHWAY(*polygon*)
2:     *roads* ← *OSM circular roads intersecting with polygon*
3:     *roadPolygons* ← *polygonize*(*roads*)
4:     **for** roadPolygon in roadPolygons **do**
5:         **if** (*roadPolygon* ⊇ *polygon*) **and**
            (*roadPolygon* ∈ *Private Roads*) **and**
            (*area*(*roadPolygon*) ≤ 1.5 ∗ *area*(*polygon*)) **then**
6:             *polygon* ← *roadPolygon*
7:         **else if** (*roadPolygon* ∈ *Private Roads*) **and**
            (*intersection*(*polygon*, *roadPolygon*) ≥
            0.2 ∗ *area*(*polygon*)) **then**
8:             *polygon* ← *Union*(*polygon*, *roadPolygon*)
9:         **else if** *roadPolygon* ∈ *Public Roads* **then**
10:            **if** *overlap*(*polygon*, *roadPolygon*) ≥
                0.5 ∗ *area*(*polygon*) **then**
11:                *polygon* ←
                  *intersection*(*polygon*, *roadPolygon*)
12:            **else**
13:                *polygon* ← *polygon* − *roadPolygon*
14:     **return** *polygon*                                                                      ▷ The corrected polygon

The basic idea is to correct the algorithmic polygons using intersecting building footprint polygons on the OSM database. Note that a polygon with $N$ vertices is specified as a sequence of (*lat*, *lng*) pairs $[(lat_1, lng_1), \cdots, (lat_N, lng_N), (lat_{N+1}, lng_{N+1})]$, where $(lat_1, lng_1) = (lat_{N+1}, lng_{N+1})$. However, the OSM polygons are sometimes incorrectly specified as a wrong sequence of (*lat*, *lng*) pairs. To account for such cases, we take the envelope of the sequence of (*lat*, *lng*) pairs using $\alpha$-shape [10].

The algorithmic polygons leak onto public roads in cases where customers mark their locations outside the gate of the PoI or due to noise in the customer locations or points on either side of a road getting clustered together. Therefore, in cases where a public road cuts across an algorithmic polygon, we retain only the largest polygon on either side of the public road (prune_polygon_via_highway presented in Algorithm 1). It must be noted however that there is no foolproof classification of private and public roads on OSM, and since the data is crowdsourced they are also error-prone. Therefore, we split the polygon cutting across a road (marked public or not) if the road is substantially long (line 6, Algorithm 1). In the case of a large PoI, internal roads form a closed loop around the PoI and such a loop is better representative of the boundary of the PoI. Therefore, we also make use of "closed private roads" to correct the boundaries of the algorithmic polygons (correct_polygon_via_highway presented in Algorithm 2). The highway tags on OSM used as private and public road proxies are specified in Table 4.

As shown in Fig. 4, the median area recall metric improves substantially to 70% with a degradation in the median area precision which is equal to 69%, and the median F-score is given by 0.69. A similar substantial improvement is also observed with OSM polygon correction applied on the baseline polygons as shown in Table 3. However, note from the last row in Table 3 that 45% of the proposed algorithmic polygons are not present on OSM.

Table 4. Highway tag proxies for public and private roads on OSM.

| Road | Highway Tag Proxy |
|---|---|
| Private | 'service', 'unclassified', 'footway', 'tertiary', 'path', 'pedestrian', 'track' |
| Public | 'primary', 'secondary', 'trunk', 'motorway', 'primary_link', 'secondary_link', 'trunk_link', 'motorway_link', 'raceway', 'bridleway', 'escape', 'bus_guideway' |

## 7 DISCUSSION

The apparent degradation in precision after OSM based polygon correction happens mainly due to ground truth polygons having marginal overlap with the neighbouring OSM building footprint polygons (belonging to a different PoI). An example is presented in Appendix B. This happens because the manually validated polygons are fit on satellite imagery taken at oblique angles and even the OSM polygons are prone to marginal errors.

Some top causes for low recall of the PoI mining algorithm are discussed below. The address locations in our dataset do not span the area of the PoI polygon. This is either due to the choice of the dataset itself (which was chosen by high order volumes per *address_id*) or due to only a few customers within the PoI being registered on the platform. The algorithm misses the inclusion of some addresses belonging to the same PoI within the homogeneous clusters. The RoBERTa embeddings are not specifically optimised for similar embeddings for addresses within the same PoIs. They are generated from a model trained as a masked language model (MLM) that predicts a token in the address given the context tokens in the same address. This is very different from the end task we are looking at. Though this learns contextual embeddings, the context is not necessarily tied up to predicting PoIs. In other words, the address embeddings for the addresses within the PoIs aren't particularly optimised for the choice of cosine distance metric used in Section 4.5.1. A representative example of such a case is the pre-processed address *'polycab india limited unit godrej genesis'* not included in the homogeneous cluster containing the pre-processed addresses [ *'godrej genesis building'*, *'futures first godrej genesis building ep gp opp to syndicate bank sector sa'*, *'futures first godrej genesis'* ].

Though the DBSCAN step splits up some homogeneous clusters falsely representing the PoI candidates, some valid sub-clusters belonging to the same PoI also are split up. They do not get merged in the homonymous merge step because these sub-clusters do not have a high confidence name.

## 8 CONCLUSION

Mining PoI polygons from address texts and address locations was achieved by jointly clustering the locations and the address embeddings through hierarchical clustering using the Euclidean distance metric. The PoI candidates were identified through cosine distance similarity between the address embeddings within the cluster and the centroid address embedding of the cluster. However, note that the address embeddings generated using RoBERTa are not particularly trained to optimise for Euclidean distance or cosine distance metrics between addresses within the same PoI. This paper however shows through empirical studies that such an unsupervised system design is still significantly useful and offers better coverage than the Mummidi-Krumm baseline algorithm evaluated on our internal dataset. We believe that a supervised approach where the address embeddings are trained to optimise a given distance metric for addresses within the same PoI would achieve a better recall metric and will be the focus of our future work. We also proposed an OSM building footprint based post-processing algorithm that can be used with any PoI mining algorithm. We used it to specifically improve the recall. We however note that the metrics are only directionally indicative. An exact validation of the output polygons will have to be done manually.

## A  OSM BASED POLYGON CORRECTION

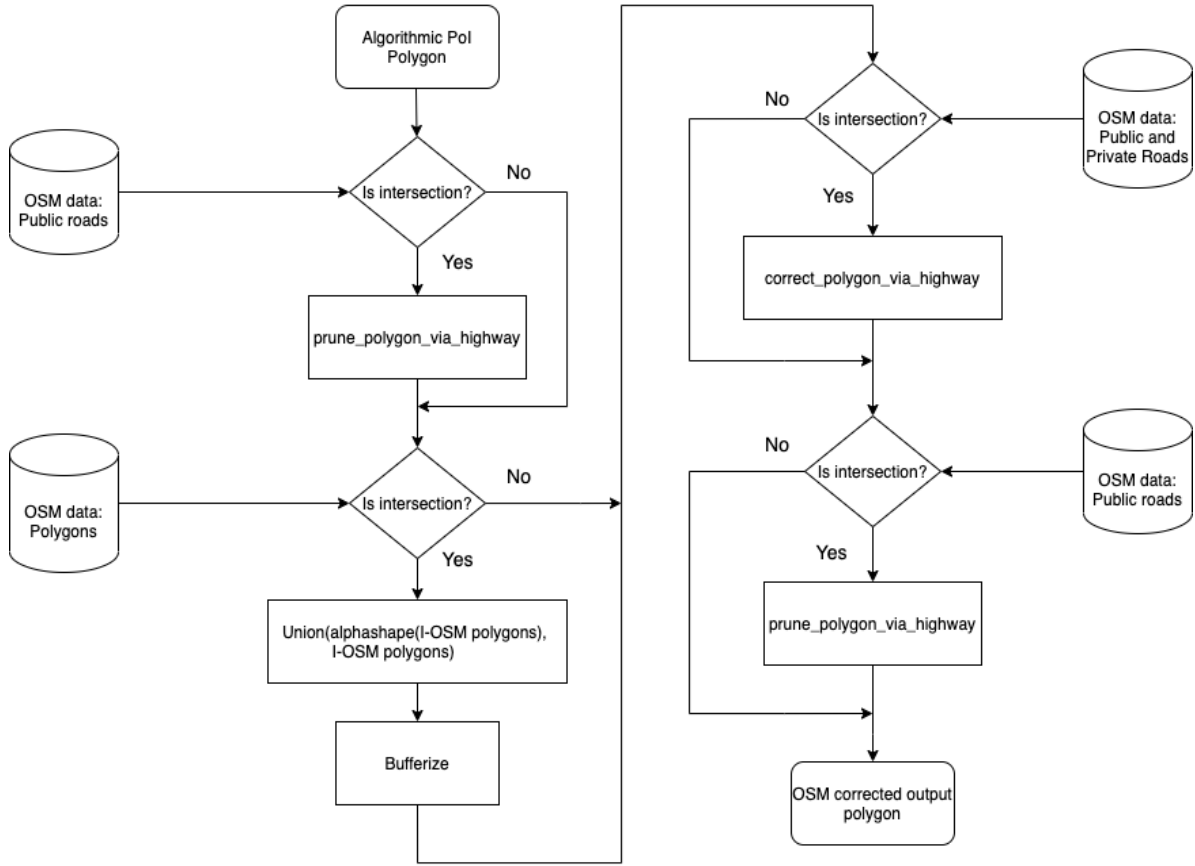

Fig. 5. Flowchart for correction of algorithmic polygons using building footprint polygons and roads from OSM database. Note that union of intersecting OSM (I-OSM) polygons in case of multiple polygons are taken and the unionised polygon is buffered by an additional 5*m* radius around the polygon to account for private space around the building footprints.

The flowchart for OSM data based algorithmic polygon correction is given in Fig. 5. Some representative examples of Algorithm 1 and 2 in action appear in Fig. 6 in the next page.

## B  PRECISION DEGRADATION

An example of an apparent degradation in precision due to marginal overlap of OSM corrected algorithmic polygon with a ground truth polygon is presented in Fig. 7 in the page after the next.

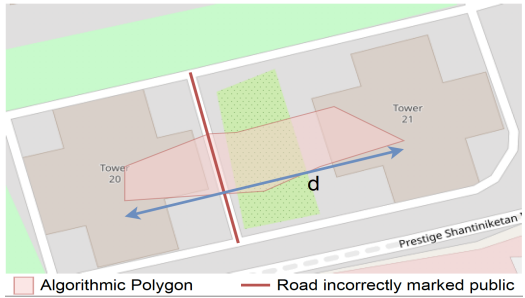

(a) Algorithm 1 (Line 6): Case of no pruning.

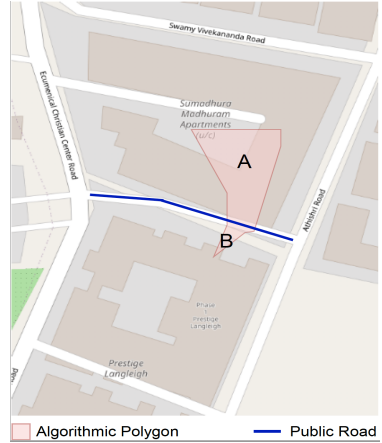

(b) Algorithm 1 (Line 6): Case of pruning.

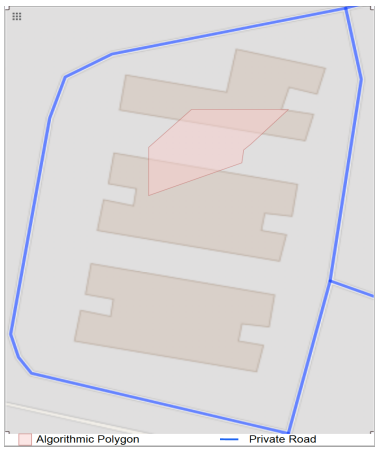

(c) Algorithm 2 (Line 5): Closed private road (blue polygon) encompassing the union of algorithmic polygon and OSM polygon. The blue polygon represents the output and the polygon encompasses the private compound of the residential complex.

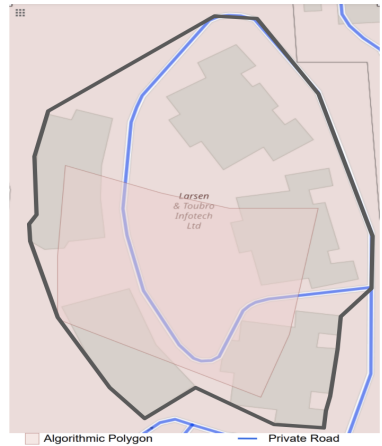

(d) Algorithm 2 (Line 7): Intersection between closed road polygon and the union of algorithmic and OSM polygons. The black polygon represents the output.

Fig. 6. Representative examples of Algorithm 1 and Algorithm 2 in action.

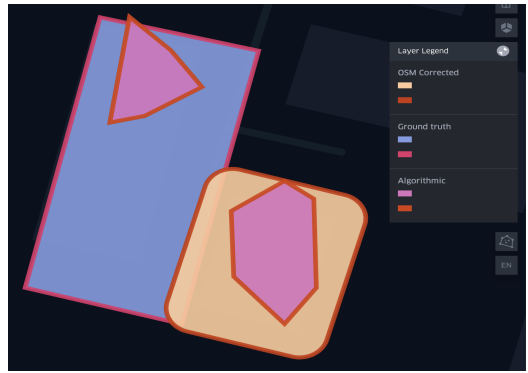

Fig. 7. The algorithmic polygon on the right (filled with pink color) does not intersect with the ground truth polygon on the left (filled with blue color) while the OSM corrected polygon (filled with brown color) intersects marginally with the ground truth polygon.